\pdfoutput=1

\documentclass[11pt]{article}

\usepackage[]{EMNLP2022}

\usepackage{times}
\usepackage{latexsym}
 \usepackage{mathrsfs}
 \usepackage{graphicx}
\usepackage{bbding}
\usepackage{amsmath}
\usepackage{CJKutf8}
\usepackage{color}

\usepackage[T1]{fontenc}

\usepackage[utf8]{inputenc}

\usepackage{microtype}

\usepackage{inconsolata}

\title{Textual Enhanced Contrastive Learning for Solving Math Word Problems}

\author{Yibin Shen\thanks{\quad This denotes equal contribution.}$\ \,^1$, Qianying Liu\footnotemark[1]$\ \,^{2}$, Zhuoyuan Mao$^2$, Fei Cheng$^2$ and Sadao Kurohashi$^2$\\
$^1$ Meituan\\
$^2$ Graduate School of Informatics, Kyoto University \\
  {\tt
  shenyibin@meituan.com;}
  {\tt
  \{ying,zhuoyuanmao\}@nlp.ist.i.kyoto-u.ac.jp;}\\{\tt \{feicheng, kuro\}@i.kyoto-u.ac.jp} \\
  }
\begin{document}
\maketitle
\begin{abstract}
Solving math word problems is the task that analyses the relation of quantities and requires an accurate understanding of contextual natural language information. 
Recent studies show that current models rely on shallow heuristics to predict solutions and could be easily misled by small textual perturbations.
To address this problem, we propose a Textual Enhanced Contrastive Learning framework, which enforces the models to distinguish semantically similar examples while holding different mathematical logic. We adopt a self-supervised manner strategy to enrich examples with subtle textual variance by textual reordering or problem re-construction. We then retrieve the hardest to differentiate samples from both equation and textual perspectives and guide the model to learn their representations.
Experimental results show that our method achieves state-of-the-art on both widely used benchmark datasets and also exquisitely designed challenge datasets in English and Chinese. \footnote{Our code and data is available at \url{https://github.com/yiyunya/Textual_CL_MWP}}

\end{abstract}

\section{Introduction}

\begin{figure}[t]
  \centering
  \includegraphics[width=\linewidth]{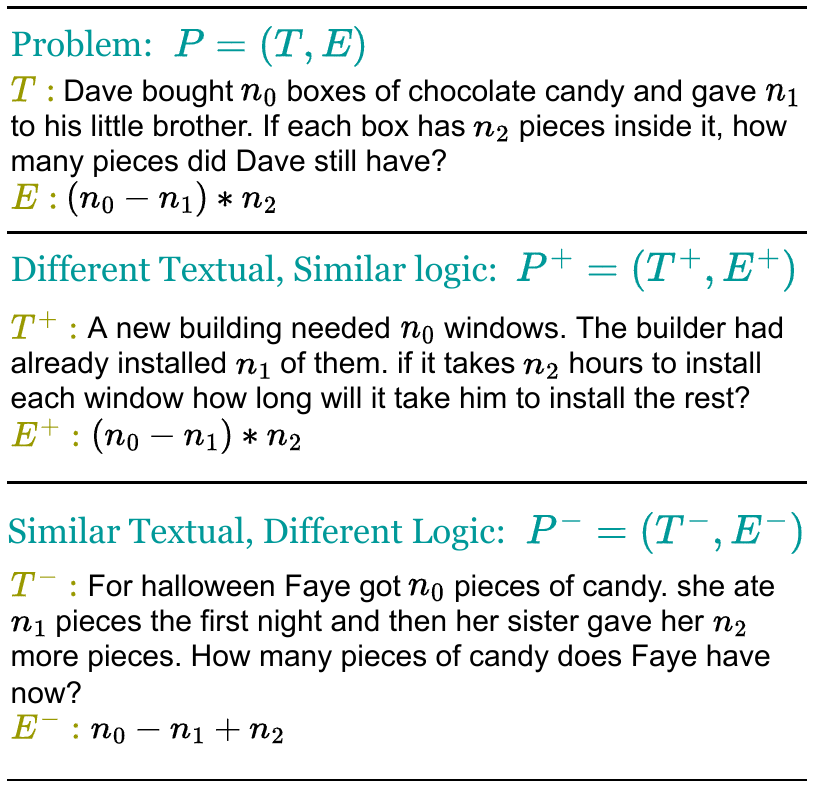}
  \caption{Example of positive data point $P^+=(T^+, E^+)$ and negative data point $P^-=(T^-, E^-)$ for an anchor $P=(T,E)$.}
  \label{fig:ex}
  \end{figure}

Solving Math Word Problems (MWPs) is the task of automatically performing logical inference and generating a mathematical solution from a natural language described math problem.
Solving MWPs is a challenging task that cannot rely on shallow keyword matching but requires a comprehensive understanding of contextual information.
For example, as shown in Figure \ref{fig:ex}, while the first problem shares high token-level overlapping with the third problem, the underlying mathematical logic is different. While on the other hand, the first and second problems have very low similarity at the textual level, while the equation solution is the same. 
The challenge of the task is that the underlying mathematical logic would change even with minor modifications in the text.
While neural network based models have greatly boosted the performance on benchmarks datasets, \citet{DBLP:conf/naacl/PatelBG21} argued that state-of-the-art (SOTA) models use shallow heuristics to solve a majority of word problems, and struggle to solve challenge sets that have only small textual variations between examples. 

Motivated by recent progress in contrastive learning methods, which is a flexible framework that has been successfully employed to representation learning in various fields~\cite{chopra2005learning, DBLP:journals/corr/abs-2005-12766,DBLP:conf/emnlp/GaoYC21}, we propose Textual Enhanced Contrastive Learning, which is an end-to-end framework that uses both textual and mathematical logic information to build effective representations. For each \textit{anchor} data point,
we find the \textit{hard example} triplet pair, which consists of a textual-different but logic-similar \textit{positive} data point $P^+$, and a textual-similar but logic-different \textit{negative} data point $P^-$. 
Our method aims to learn an embedding space where the vector representations of $P$ and $P^+$ in Figure \ref{fig:ex} are mapped close together, since they hold the same mathematical logic even though the textual expression is entirely different; on the other hand, because $P$ and $P^-$ have similar textual expression but different mathematical logic, their vector representations could be separated apart.

To build such triplet pairs, we use a retrieval-based method to search in the training data.
We consider the equation annotation as the representation of the mathematical logic in the example, and retrieve a positive and negative bag of data points according to equation similarity.
Then we further use textual similarity to choose the \textit{hard examples} in the bags, where positive examples have low textual similarity with the anchor and vice versa. Given such \textit{hard sample} data, Contrastive Learning could empower the representations by leading the model to distinguish these potential disorienting examples in the training stage. 

Such approaches to retrieving triplet pairs from human-annotated training data via label annotation are considered as \textit{supervised} contrastive learning. Another research line of contrastive learning is \textit{self-supervised} contrastive learning, which does not require labeled data and use data augmentation methods to generate the positive or negative data points~\cite{DBLP:conf/icml/ChenK0H20, DBLP:conf/cvpr/He0WXG20, DBLP:conf/nips/GrillSATRBDPGAP20}. In the task of solving MWPs, we can leverage \textit{self-supervised} supervision by generating new examples via performing synchronized changes to text and equations. The generated data is naturally \textit{hard sample} data, because the textual expression is similar to the origin example, while the equation could be either changed or the same. Specifically, we leverage Reversed Operation-based Data Augmentation \cite{liu2021roda} and a Question Reordering-based augmentation to form new data points. 
By enhancing the model to detect the small perturbations in the augmented examples, contrastive learning forces the model to learn more effective representations of contextual information.

While previous studies also used Contrastive Learning to improve representations for solving MWPs \cite{li2022seeking}, their method is limited to supervised contrastive learning, ignores textual information during constructing the contrastive learning pairs and requires two step pre-training and re-training. Our method pushes the model to learn better text representations and understand the most minor textual variance from these textual enhanced \textit{hard samples} from both \textit{supervised} and \textit{self-supervised} perspectives.

We conduct experiments on two widely used datasets, the English dataset ASdiv-A~\cite{DBLP:conf/acl/MiaoLS20} and the Chinese dataset Math23K~\cite{DBLP:conf/emnlp/WangLS17}. To further investigate how our method improves the ability of the model to detect small textual perturbations, we collect a Chinese challenge set Hard Example (HE)-MWP. We perform experiments on two challenge sets of MWPs, the English Asdiv-Adv-SP dataset~\cite{DBLP:conf/emnlp/KumarMP21} and the Chinese HE-MWP dataset. Experimental results show that our method achieves consistent gains under different languages and settings, demonstrating the effectiveness of our method.

\section{Related Work}

\subsection{Solving Math Word Problems}

There are various research lines in solving math word problems. Early studies majorly rely on rule-based methods \cite{Bobrow:1964:NLI:889266,Charniak:1969:CSC:1624562.1624593}. Statistical machine learning methods were developed to map math word problems to specific equation templates~\cite{kushman2014learning,roy-roth-2015-solving,koncel2015parsing,roy2017unit}. Another research line uses semantic parsing-based methods to transform the input text into structured representations that could be parsed to obtain the answer~\cite{roy2018mapping,shi2015automatically,zou2019text2math}. Recent studies focus on using a sequence-to-sequence (seq2seq) framework that takes in the text descriptions of the MWPs and predicts the answer equation. To improve the framework, various studies have investigated task designing task specialized encoder and decoder architectures~\cite{wang2018mathdqn, wang2019template, xie2019goal, liu-etal-2019-tree, guan-etal-2019-improved, zhang2020graph, ijcai2020-555, shen-jin-2020-solving}, using pre-trained models~\cite{tan2021investigating,liang2021mwpbert} and leveraging auxiliary tasks~\cite{liu2021roda, shen2021generate, li2022seeking, shen2022seeking}. Various auxiliary tasks have been introduced to improve model performance. \citet{shen2021generate} introduced a reranking loss that reranks the beam search predictions. \citet{huang2021recall} introduced a memory augmented subtask that gives guidance during the decoding stage. The closest study to our research is \cite{li2022seeking}, which uses equations as searching schema to build positive-negative pairs, and then perform contrastive learning. However, their research ignores textual information while building contrastive learning triplet pairs and limits supervised contrastive learning.

MWP solvers have achieved relatively high performance on benchmark datasets. However, the extent to which these solvers truly understand language and numbers remains unclear. Various studies either use data augmentation to help the model improve robustness and performance on hard cases or develop adversarial examples and challenge sets to evaluate the robustness of MWP solvers against textual variance. \citet{liu2021roda} proposed a data augmentation method that reverses the mathematical logic in the problem to generate a new example. \citet{DBLP:conf/naacl/PatelBG21} constructed a challenge set of the math word problem in which the problem text only has a small variance.  \citet{DBLP:conf/emnlp/KumarMP21} investigated adversarial attack on MWP solvers. The challenge sets and adversarial attacks show that current MWP solvers use shallow
heuristics to solve a majority of word problems and fail to detect subtle textual variance. 

\subsection{Contrastive Learning}

Contrastive Learning was first adopted in Computer Vision to learn representations of images via self-supervision without human annotation~\cite{DBLP:conf/icml/ChenK0H20, DBLP:conf/cvpr/He0WXG20, DBLP:conf/nips/GrillSATRBDPGAP20}. Self-supervised contrastive learning is applied in NLP to learn sentence representations. Back translation \cite{DBLP:journals/corr/abs-2005-12766} and dropout \cite{DBLP:conf/emnlp/GaoYC21} are used to construct positive-negative contrastive learning triplets. These perturbation-based techniques are not suitable for MWP solvers, that MWPs are sensitive to small textual variance and the perturbation might introduce noise.

\citet{DBLP:conf/nips/KhoslaTWSTIMLK20} first introduced supervised contrastive learning in Computer Vision by modifying the loss to allow supervision from label annotations. In NLP, various studies have introduced natural language inference (NLI) datasets as supervised annotations for contrastive learning~\cite{DBLP:conf/emnlp/ReimersG19, DBLP:conf/emnlp/GaoYC21}. The agreement of equation annotations of MWPs can be considered a form of NLI, that our supervised contrastive learning could be considered a transformation of these methods.

\begin{figure*}[t]
  \centering
  \includegraphics[width=\linewidth]{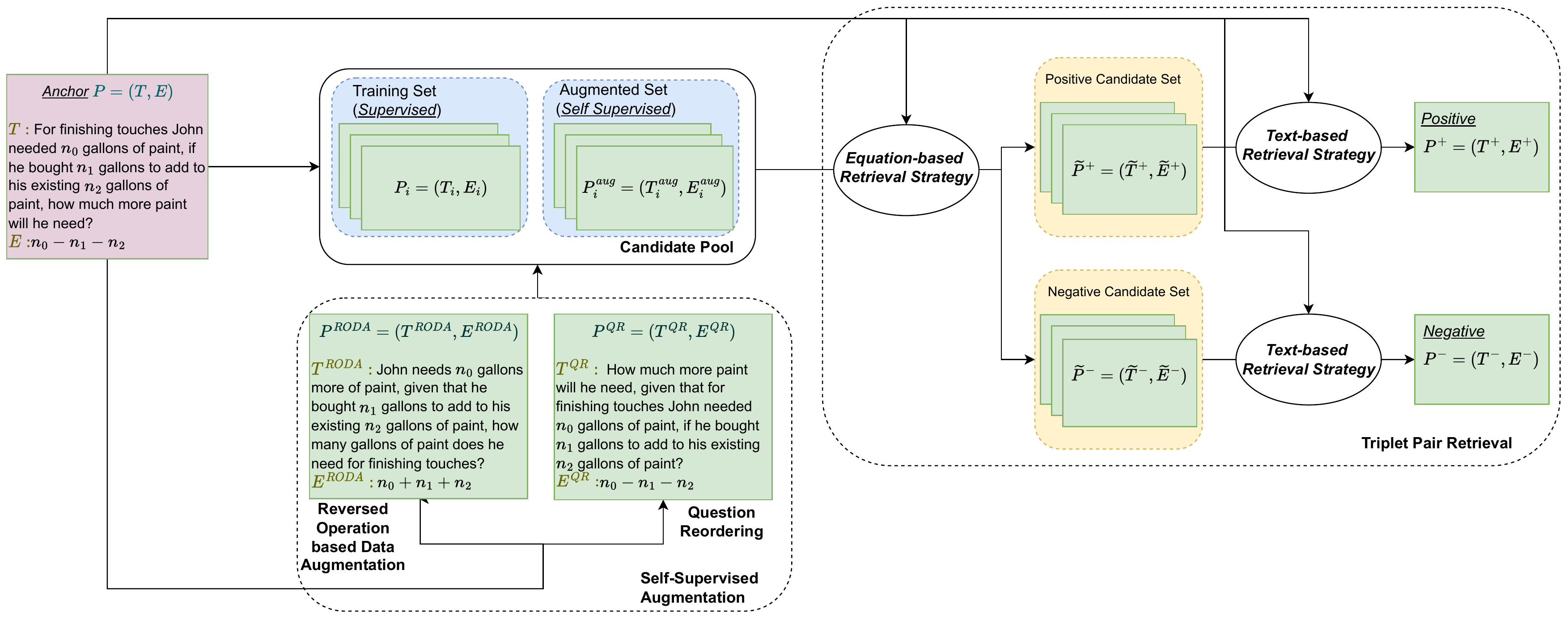}
  \caption{Overview of the contrastive learning triplet pairs retrieval procedure. }
  \label{fig:model}
  \end{figure*}


\section{Methodology}

We use Contrastive Learning to obtain text features with high differentiation of small perturbations, so that for each \textit{anchor} data point $P=(T,E)$, where $T$ stands for the text and $E$ stands for the equation, we construct a pair of examples 
\textit{positive} data point $P^+=(T^+, E^+)$ and \textit{negative} data point $P^-=(T^-, E^-)$, and then use contrastive learning loss to map the representation of $P$ and $P^+$ closer and vice versa. The pipeline of the triplet pairs retrieval is shown in Figure \ref{fig:model}. 
We first construct a candidate pool, which consists of \textit{supervised} training data $\{P_i\}$ and augmented \textit{self-supervised} data $\{P^{aug}_i\}$ as shown in the blue part of Figure \ref{fig:model}. The \textit{self-supervised} data is generated by two methods, Reversed Operation based Data Augmentation (RODA) and Question Reordering (QR), which is explained in Section \ref{ss:aug}. Then we perform two-step retrieval to retrieve the triplet pairs as described in Section \ref{ss:ret}. We first use an equation-based retrieval strategy to extract \textit{positive} candidate set $\{\widetilde{P}^+\}$ and \textit{negative} candidate set $\{\widetilde{P}^-\}$, and then further introduce textual information by choosing one example from the candidate set via a text-based retrieval strategy. Finally, we train the MWP solving model that maps $T$ to $E$ by considering both the contrastive learning and solution equation generation objective, as described in Section \ref{ss:model}.

\subsection{Enriching Candidate Pool via Self-Supervised Augmentation}
\label{ss:aug}
The \textit{self-supervised} examples are challenging for the model to distinguish; while the perturbation in the text expression is extremely subtle, the corresponding mathematical logic could still change. Compared to the supervised examples that are retrieved from the training data, these \textit{self-supervised} samples place a higher demand on the model's ability to detect subtle changes and understand contextual information. We generate task-orientated augmented examples from training set data point $P=(T,E)$ via two methods that obtain reliable new text-equation examples by modifying the text and equation in the same logic at the same time. We split the sentences with punctuation marks to a question followed by various declarative sentences $T=\{S_1, S_2,..., S_{k-1}, Q_k\}$. The question sentence is always the last sentence for Asdiv, and we check whether interrogative pronouns are in the last sentence for Math23K. 


\subsubsection{Question Reordering}

We move the question to the front of the MWP to form a reordered new MWP similar to \citet{DBLP:conf/emnlp/KumarMP21}. Given a problem text $T=\{S_1, S_2,..., S_{k-1}, Q_k\}$, we move the question $Q_k$ to the front of the problem text to form a new problem text $T^{QR}=\{Q_k, S_1, ..., S_{k-1}\}$ while the rest of the text remains the same. We simultaneously edit the equation $E^{QR}$ so that the variables match with the new text order. The new example $P^{QR}=(T^{QR},E^{QR})$ could either be a positive example that holds the same equation as $P$ or a negative example that holds a different equation since the variable order might change during the reordering. The high textual similarity but rotated variable order pushes the model to learn representations that can differ from these small textual perturbations.

\subsubsection{Reversed Operation based Data Augmentation}

We perform RODA~\cite{liu2021roda} that generates a new example by asking a question about one of the original given variable. Given a problem text $T=\{S_1, S_2,..., S_{k-1}, Q_k\}$ where the question $Q$ asks about an unknown variable $n_{ans}$, RODA chooses a known variable $n$ in one of the declarative sentence $S_i$, and then generates a problem text which asks about this variable. To generate such an example, $S_i$ is transformed to a question $Q_{S_i}$ which asks a question of $n$, while $Q$ is transformed to a declarative sentence $S_k$ describing $n_{ans}$. We reorder the problem text by swapping the two sentences, that a new problem text $T^{RODA}=\{S_1, ... S_k, .. S_{k-1}, Q_{i}\}$ is generated. Simultaneously we edit the equation by resolving the equation expression $E^{RODA}$ of $n$ given $n_{ans}$. While $P^{RODA}=(T^{RODA}, E^{RODA})$ has a very similar textual description of $P$, the underlying equation could be completely different, which could benefit the model via contrastive learning. RODA requires text parsing and transformation rules to modify the text and equation. For Chinese, it can cover 93\% of the examples, and for English, it covers 60\% of the examples. The generated text has a 0.83 out of 1 coherent score reported by human evaluation by \citet{liu2021roda}.

\subsection{Triplet Pair Retrieval}
\label{ss:ret}

We construct the positive and negative triplet pairs from both textual and logical perspectives. For a given problem $P$, the positive sample $P^+$ is considered to be a problem with similar equation expressions but relatively different text descriptions; the negative sample $P^-$ is considered to be a problem with highly textual similarity but different equation expression. However, it requires a time-consuming bruce-forth enumeration of all possible example pairs to find such optimal positive and negative samples. 
Considering the computational complexity, we break down the retrieval to a two-step pipeline. we adopt a heuristic searching algorithm to construct positive and negative samples $(P^+, P^-)$ as follows:

\begin{enumerate}
    \item Construct a similarity matrix $M$ of all equation expressions $\{E_1, E_2, ...E_n\}$ in the training set, where $M_{ij}$ is the similarity of equation expression $E_i$, $E_j$. 
    \item For a given anchor $P$, Retrieve a \textit{positive} candidate set $\{\widetilde{P}^+\}$ and a \textit{negative} candidate set $\{\widetilde{P}^-\}$ of samples from the training set of the data via equation expression similarity.
    \item Extract the best \textit{positive} example $P^+$ and the best \textit{negative} example $P^-$ via textual similarity.
\end{enumerate}

We investigate various strategies to retrieve $(P^+, P^-)$ from both equation-based and text-based perspectives.

\subsubsection{Equation-based Retrieval Strategy}

To evaluate the equation similarity during the retrieval, we design an equation similarity metric $Sim_{eq}$ based on length-wise normalized tree edit distance (TED). TED is defined as the minimum-cost sequence of node operations that transform one tree into another and is a well-known distance measure for hierarchical data. We define the TED of two equation expressions $E_1, E_2$ as the TED of their abstract syntax tree. The similarity of two equation expressions $E_1, E_2$ is defined as:

\begin{equation*}
    Sim_{eq}(E_1, E_2) = 1 - \frac{TED(E_1, E_2)}{ |E_1| + |E_2| }
\end{equation*}

Given this equation similarity metric, we design two retrieval strategies.

\paragraph{Exact Match} The \textit{positive} candidate set $\{\widetilde{P}^+\}$ is constructed of the examples that meets $Sim_{eq}(E, E_i) = 1$, which means their equation expression satisfies $E = E_j$. If only the \textit{anchor} itself holds this equation expression, the \textit{positive} candidate set $\{\widetilde{P}^+\}$ has only the \textit{anchor} $P$. The \textit{negative} candidate set $\{\widetilde{P}^-\}$ is constructed of the examples that meets $argmax_{E_i \neq E}(Sim(E, E_i))$, which holds the closest equation considering the \textit{anchor}.

\paragraph{Nearest Neighbour} The \textit{positive} candidate set is constructed of the examples that meets $argmax_{E_i, T_i \neq T}(Sim_{eq}(E, E_i))$. If no other example holds the same equation expression as the \textit{anchor}, the \textit{positive} candidate set $\{\widetilde{P}^+\}$ takes the examples that has the nearest neighbour equation expression. The \textit{negative} candidate set $\{\widetilde{P}^-\}$ is constructed of the examples that meets $argmax_{E_i \neq E^+}(Sim_{eq}(E, E_i))$, which holds the closest equation considering the \textit{positive} example.

The \textit{positive} and \textit{negative} candidate sets are then further screened by the text-based strategy.

\subsubsection{Text-based Retrieval Strategy}

To lead the model to differentiate mathematical logic from similar textual expressions, we use textual-based information to select the $(P^+, P^-)$ pair. 
We select the \textbf{lowest} textual similarity score example from the positive candidate set $\{\widetilde{P}^+\}$, which is the example with different textual expression but the same mathematical logic; and select the \textbf{highest} textual similarity score example from the negative candidate set $\{\widetilde{P}^-\}$, which is the example with similar textual expression but different mathematical logic. We design two similarity measurement metrics for this stage.

\paragraph{BERTSim} Sentence-BERT (SBERT) is a strong sentence representation baseline model~\cite{DBLP:conf/emnlp/ReimersG19}. We calculate the cosine similarity of the SentBERT representation of the two sentences to obtain the similarity score:

\begin{equation*}
    Sim^{BERTSim}_{text} = \frac{SBERT(T_1) \cdot SBERT(T_2)}{||SBERT(T_1)|| ||SBERT(T_2)||}
\end{equation*}

The value range of $Sim^{BERTSim}_{text}$ is from $[-1,1]$.

\paragraph{Bi-direction BLEU} BLEU is a widely used evaluation metric for text generation that measures the similarity between the generated text and the reference. We design a Bi-direction BLEU since BLEU is a not symmetrical similarity metric, which is defined as:

\begin{equation*}
    Sim^{BiBLEU}_{text} = \frac{BLEU(T_1, T_2) + BLEU(T_2,T_1)}{2}
\end{equation*}

The value range of $Sim^{BiBLEU}_{text}$ is from $[0,1]$.

\begin{table*}
\centering
\begin{tabular}{lcccc}
\hline
\textbf{Dataset} & Math23K & Asdiv-A & HE-MWP & Adv-Asdiv  \\
\hline
Language&zh&en&zh&en\\
Domain&general&general&challenge&challenge\\
\#Train&21,162&1,218&-&-\\
\#Dev/\#Test&1,000 / 1,000&- / -&- / 400&- / 239\\
\#Equation Templates & 3,104
&66&231&66\\
\hline
\end{tabular}
\caption{Statistics and details of the datasets.}
\label{tab:dataset}
\end{table*}
\subsection{Training Procedure}
\label{ss:model}
\begin{figure}[t]
  \centering
  \includegraphics[width=\linewidth]{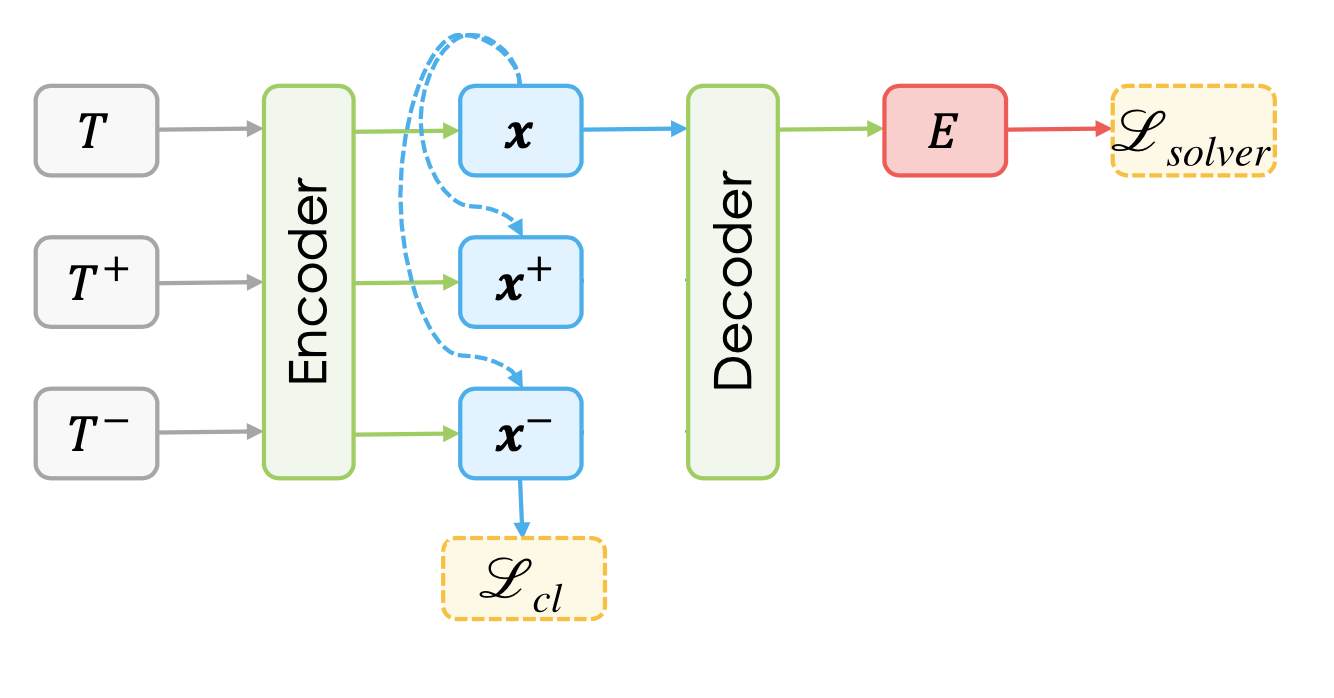}
  \caption{Overview of the training procedure. }
  \label{fig:train}
  \end{figure}

We show the training procedure in Figure \ref{fig:train}. The training loss consists of the MWP solving loss $\mathcal{L}_{solver}$ and the contrastive learning loss $\mathcal{L}_{cl}$. 

\paragraph{MWP Solving Model} We follow \citet{li2022seeking} and use the strong baseline model BERT-GTS as MWP solving model. The pre-trained language model BERT, which provides strong textual representations, is leveraged as the encoder. For the decoder, we use Goal-driven tree-structured MWP solver (GTS) 
\cite{xie2019goal}.
 GTS directly generates the prefix notation of the solution equation by using a recursive neural network to encode subtrees based on the representations of its children nodes with the gate mechanism. With the subtree representations, this model can well structured information of the generated part to predict a new token.

\paragraph{Contrastive learning} Contrastive learning is performed on triplets pairs $(P, P^+, P^-)$ by pulling the representations of $T$ and $T^+$ together and pushing apart the representations of $T$ and $T^-$. We follow the contrastive learning framework in \citet{DBLP:conf/icml/ChenK0H20}, which takes an in-batch cross-entropy objective. Let $x_i$ denote the encoder representation of $P$, the training objective for $(x_i, x_i^+, x_i^-)$ within the batch of $N$ triplet pairs is:

\begin{align*}
         &\mathcal{L}_{cl}=\\
     &-\frac{1}{N}\sum^N_{i=1}log\frac{e^{cos(x_i, x_i^+)/\tau}}{\sum^N_{j=1}(e^{cos(x_i, x_j^+)/\tau}+e^{cos(x_i, x_j^-)/\tau})}
\end{align*}

where $\tau$ is the temperature hyperparameter. 

Assume the prediction target equation of $P$ is $y$, the final training objective is to minimize the sum of the MWP solution equation generation negative log likelihood loss $\mathcal{L}_{solver}$ and the contrastive learning loss $\mathcal{L}_{cl}$:

\begin{align*}
    \mathcal{L} &= \mathcal{L}_{solver} + \alpha * \mathcal{L}_{cl} 
\end{align*}

\begin{table*}
\centering
\begin{tabular}{lcccccc}
\hline
\textbf{Model} &Cand. Pool &Math23K & Asdiv-A & HE-MWP &  Adv-Asdiv-SP  \\
\hline
GTS~\cite{xie2019goal}&-&75.6&68.5&-&21.2\\
G2T~\cite{zhang2020graph}&-&77.4&71.0&-&23.8\\
pattern CL~\cite{li2022seeking}&train&83.2&-&-&-\\
\hline
BERT-GTS&-&82.9&73.4&55.5&59.9\\
w/ \textit{supervised CL}&train&84.1&74.2&57.2&63.7\\
w/ RODA CL&RODA+train&84.3&74.3&64.1&64.1\\
w/ QR CL&QR+train&84.2&74.4&62.5&66.2\\
w/ CL&RODA+QR+train&85.0&74.6&69.5&66.9\\
\hline
\end{tabular}
\caption{Results on MWP datasets. All experiments only compute MWP solving loss on the training set. The candidate pool only affects the choice of positive and negative examples in the CL loss.}
\label{tab:result}
\end{table*}

\section{Experiments}

\subsection{Datasets}

We perform experiments on four datasets, including two widely used datasets to verify the generalization ability of our method and two challenge test sets to show further how our method can enhance the robustness of the model. We show detailed statistics of the datasets in Table \ref{tab:dataset}.

\paragraph{Math23K} is a Chinese dataset that contains 23,161 math word problems of elementary school level~\cite{DBLP:conf/emnlp/WangLS17}. We use the standard train-test split setting of this dataset for the experiment.

\paragraph{Asdiv-A} is the the arithmetic subset of ASDiv which has 1,218 MWPs mostly up to grade level 4~\cite{DBLP:conf/acl/MiaoLS20}. Experiments of this dataset are evaluated by 5-cross validation.

\paragraph{HE-MWP} Since no challenge dataset has been developed for Chinese MWP solving, and existing challenge datasets have limited types of equation templates, we use RODA and QR on Math23K validation set to generate examples that are semantically similar to the original input but deceive the model into generating an incorrect prediction. We randomly sample a subset of 600 examples from the RODA result of the development set of Math23K and then manually delete the examples that the text is not coherent. Then we randomly select 400 examples out of this cleaned subset.

\paragraph{Adv-Asdiv-SP} is challenge set of Asdiv-A, which is constructed of adversarial examples \cite{DBLP:conf/emnlp/KumarMP21}. These adversarial examples are generated by sentence paraphrasing. 

Results of the challenge datasets are tested on the highest performance models trained on the corresponding benchmark datasets.

There exists other MWP datasets, which are relatively less challenging such as ALG514, DRAW1K and MAWPS \cite{kushman2014learning, upadhyay2017annotating,koncel2016mawps}, or noisy such as Dolphin18K~\cite{huang2016well} or use semantic parsing as annotation such as MathQA~\cite{amini2019mathqa}. We use the two benchmarks Math23K and Asdiv-A because they are both clean and challenging with mathematical equation annotations.

\subsection{Implementation Details}

We use two language-specific BERT-base models as the problem encoder\footnote{English: https://huggingface.co/bert-base-uncased, Chinese: https://huggingface.co/yechen/bert-base-chinese }. For both models, the maximum text length of the encoder is fixed at 256, and the maximum equation generation length of the decoder is fixed at 45. The decoder 
embedding size is 128. The batch size is 16, with learning rate of 5e-5. We tune the hyperparameters temperature $\tau$ in the set of \{0.05, 0.1, 0.2\} and $\alpha$ in the range [0.1, 0.9]. Experiments of the Chinese datasets are conducted on V100 and RTX 3090 with approximately 6 hours of runtime. Experiments of the English datasets are conducted on 1080Ti with approximately 1-hour runtime.

\section{Results and Analysis}

\begin{table}
\centering
\begin{tabular}{l|cc}
\hline
&\multicolumn{2}{c}{\textbf{Eq Strategies}}\\
\hline
\textbf{Text Strategies} & EM & NN \\
\hline
Random &83.2&82.3\\
BERTSim&83.6&83.1\\
Bi-BLEU&\textbf{84.1}&83.2\\
\hline
BERT-GTS & 82.9 &\\
\hline
\end{tabular}
\caption{Results of different retrieval strategies for supervised contrastive learning.  EM denotes exact match. NN denotes nearest neighbour. Random denotes randomly choosing an example from the candidate set. BERTSim and Bi-BLEU denotes choosing the examples by similarity metric.}
\label{tab:strategy}
\end{table}

\subsection{Pre-examination on Retrieval Strategy}

We conduct a breakdown analysis on the most complex dataset Math23K of different retrieval strategies. We investigate the performance of different retrieval strategies for supervised contrastive learning. As shown in Table \ref{tab:strategy}, for the equation-based retrieval strategy, the exact match equation strategy is more effective than the nearest neighbour strategy. This shows that the positive sample for the anchor must have accurate same mathematical logic for contrastive learning to benefit the performance. Both text-based retrieval strategies can improve the MWP solving performance compared to the random choosing baseline, demonstrating the effectiveness of introducing textual information for contrastive training. With textual-based retrieval, the extracted positive and negative examples would form \textit{hard examples} that can push the model to differ textual-similar but logic different examples. Bi-BLEU also has a slightly higher performance than BERTSim. In the following experiments, we use the best combination of EM and Bi-BLEU as retrieval strategies.

\subsection{Main Results}

We show the results of our method compared with other baselines in Table \ref{tab:result}. In addition to our baseline BERT-GTS model, we also investigate three strong baseline models. \textbf{GTS} \cite{xie2019goal} uses an LSTM encoder and the same decoder as BERT-GTS that generates the abstract syntax trees through a tree structure decoder in a goal-driven manner. \textbf{G2T} \cite{zhang2020graph} is a graph-to-tree model that uses a graph-based encoder for representing the relationships and order information among the quantities. \textbf{Pattern CL} \cite{li2022seeking} proposes a pattern-based contrastive learning, that considers the equation similarity with supervised contrastive learning. We can see from the results that our method outperforms previous studies in all datasets. Compared to \textbf{Pattern CL} which ignores textual information, our method allows the model to have a stronger ability to bridge text descriptions to mathematical logic even using the same candidate pool. The \textit{self-supervised} methods outperform the \textit{supervised} settings, especially on challenge datasets, demonstrating the effectiveness of leading the model to learn contextual representations of small textual perturbations.

On benchmark datasets, we achieve 2.1\% points of improvement on Math23K and 1.2\% points of improvement on Asdiv-A. One major reason is that RODA can only generate 394 examples for the English dataset Asdiv-A. In contrast, it can generate 47,318 examples for the Chinese dataset Math23K because English has more strict grammar than Chinese. On challenge datasets, we achieve 14\% points of improvement on HE-MWP dataset and 7.0\% points of improvement on Adv-Asdiv-SP dataset. For HE-MWP ablation, RODA is more effective since it could introduce new mathematical logic examples. For Adv-Asdiv-SP, since QR is similar to paraphrasing techniques, it gains more improvement with self-supervised supervision.

\subsection{Visualization and Case Study}

\begin{figure}[t]
  \centering
  \includegraphics[width=0.48\textwidth]{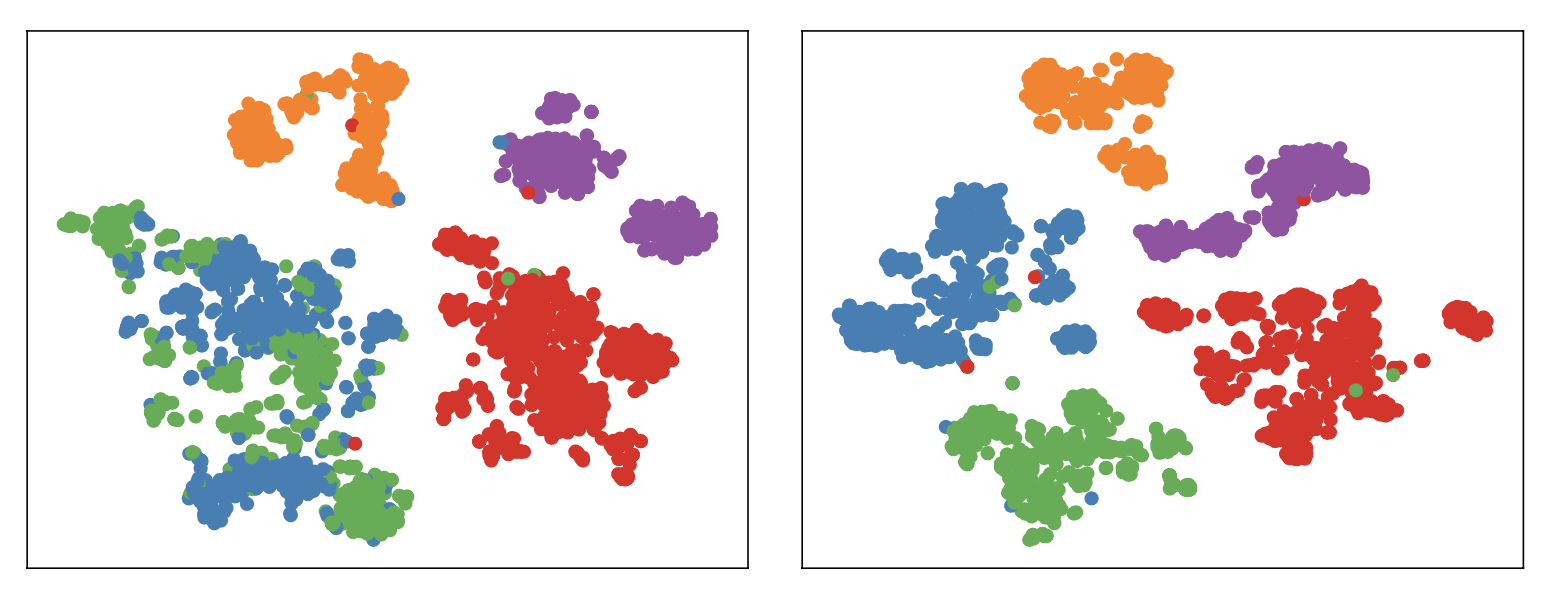}
  \caption{T-SNE Visualization results of BERT-GTS w/o (left) and w/ CL (right). }
  \label{fig:tsne}
  \end{figure}
  
  \begin{figure}[t]
  \centering
  \includegraphics[width=0.24\textwidth]{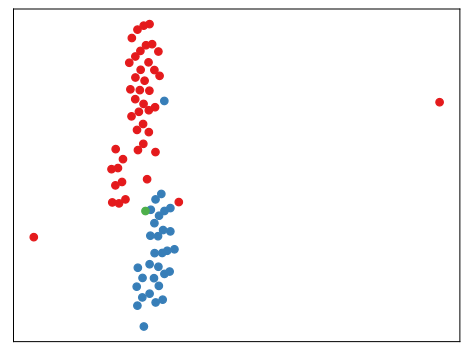}\hfill
  \includegraphics[width=0.24\textwidth]{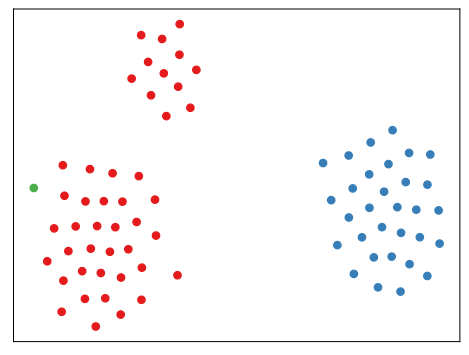}
  \caption{T-SNE visualization for the case study on BERT-GTS w/o (left) and w/ CL (right). }
  \label{fig:case}
  \end{figure}

We show T-SNE visualization results of the representations of examples from the top-five frequent equation templates in Math23K: $n_1*n_2/n_3$, $n_1*n_2$, $n_1/n_2$, $n_2/n_1$ and $n_1*(1-n_2)$, which refers to orange, red, blue, grean and purple in Figure \ref{fig:tsne}. We can see that compared to the BERT-GTS baseline on the left subfigure, in the right subfigure, the text representations of the same equations are pulled closer via our contrastive learning, and the representation of different equations are separated apart, which shows that our method can benefit the representation learning. 

We further investigate how our method improves the representation via case study.
In Table \ref{tab:case} the BERT-GTS baseline could not infer from the textual description that the side area of a cylinder is the area of a rectangle but rather uses shallow heuristics when the word "cylinder" is encountered and generates the constant $\pi$. By constructing positive and negative sample pairs from both expressions and textual descriptions and changing the representation space via contrastive learning, the model is not misled by the keywords and correctly infers that the mathematical logic is to calculate the area of a rectangle so that the model with contrastive learning generated the correct result. We also show T-SNE visualization of the representation in Figure \ref{fig:case}. The red dots are examples with the keyword rectangle and hold the equation $n_1*n_2$. The blue dots are the examples that hold the equation $\pi*n_1$ or $\pi*n_2$. The green dot is the studied case. We can see that while BERT-GTS fails to separate the representation of the case from the cylinder or circle-related equations, contrastive learning helps the model to differentiate such confusing examples, learn better representations, and predict the answer correctly.

  \begin{CJK*}{UTF8}{gbsn}
  \begin{table}
\centering
\begin{tabular}{p{40pt}p{150pt}} 
\hline

Text&\small{用一张长 $n_1$ 厘米，宽 $n_2$ 厘米的长方形纸围成一个最大 的圆柱，圆柱的侧面积为多少平方厘米?}\\
EN&\small{Given a piece of paper $n_1$ centimeters long and $n_2$ centimeters wide, How many square centimeters is the lateral area of the largest cylinder enclosed by the rectangle?}\\
w/o CL&$\pi*n_2$ (\textcolor{red}{\XSolid}) \\
w/ CL &$n_1*n_2$ (\textcolor{green}{\Checkmark})\\

\hline
\end{tabular}
\caption{Case study on Math23K example. w/o CL denotes the BERT-GTS baseline. w/ CL denotes using contrastive learning.}
\label{tab:case}
\end{table}
\end{CJK*}

\subsection{Combination with Data Augmentation}

\begin{table}
\centering
\begin{tabular}{lc}
\hline
\textbf{Model} & Acc \\
\hline

baseline & 82.9 \\
\hline
+QR aug w/o CL& 84.9 \\
+QR aug w/ CL & \textbf{85.2}\\
\hline
+RODA aug w/o CL& 84.8\\
+RODA aug w/ CL& \textbf{86.4}\\
\hline
\end{tabular}
\caption{Results of using augmented example for both training and contrastive learning.}
\label{tab:aug}
\end{table}

While the high-quality and challenging augmented examples have shown remarkable effectiveness for contrastive learning, a question remains whether contrastive learning is still effective when these augmented examples are directly used as training data. Thus, we further investigate using the augmented examples as \textit{anchors}. We use the augmented examples and the original data as training data and perform supervised contrastive learning in the training data. As shown in Table \ref{tab:aug}, we can see that while the augmented examples improve the performance, contrastive learning can further boost the performance, achieving SOTA results on Math23K.

\section{Conclusion}

In this paper, we propose a Textual Enhanced Contrastive Learning framework, which leverages both \textit{supervised} and \textit{self-supervised} supervision to help the model understand contextual information and bridge subtle textual variance to mathematical logic. We use two novel task-specific data augmentation methods to enrich the candidate pool with examples with minor textual variance for contrastive learning triplet pair retrieval. We design a two-stage retrieval method to find \textit{hard example} triplet pairs with both equation and textual information and investigate various retrieval strategies. Experimental results show that our method gained improvement on both benchmark datasets and challenge datasets in English and Chinese. We also conduct visualization for representation distribution on different equations and also on a case study, which shows our method can benefit the representation learning. With the combination of data augmentation, our method still improves the performance and achieves SOTA results on Math23K dataset.

\section*{Limitations}

While our framework extracts contrastive learning triplet pairs with light computational complexity, we observe that such a two-stage retrieval strategy might not be optimal under certain circumstances. 

We build the framework assuming that methods with similar mathematical logic (i.e., high equation similarity) would form challenging negative examples. However, especially for \textit{self-supervision}, such an assumption can block out the augmented small variance examples from consideration for triplet pairs because their equation might not be the most similar one. This is more severe when using RODA for self-supervised augment. The generated examples of RODA usually have relatively low equation similarity with the origin example. However, the RODA examples remain challenging as we can see the performance of HE-MWP still has a gap of 15\% points compared to the original Math23K datasets.

 A strategy that considers equation and textual similarity in the same stage could be introduced to fill this gap. However, such strategies cannot avoid the heavy computational complexity caused by calculating the metric of all data pairs. This could be reduced by recent studies in rapid embedding retrieval algorithms such as FAISS~\cite{johnson2019billion} by transforming the equation similarity to embedding similarity via embedding training methods. This remains as future work in this paper.

\section*{Acknowledgements}

  This work is partially supported by JST SPRING Grant No.JPMJSP2110 and KAKENHI No.22J13719, Japan.

\bibliography{anthology,custom}

\begin{thebibliography}{41}
\expandafter\ifx\csname natexlab\endcsname\relax\def\natexlab#1{#1}\fi

\bibitem[{Amini et~al.(2019)Amini, Gabriel, Lin, Koncel-Kedziorski, Choi, and
  Hajishirzi}]{amini2019mathqa}
Aida Amini, Saadia Gabriel, Shanchuan Lin, Rik Koncel-Kedziorski, Yejin Choi,
  and Hannaneh Hajishirzi. 2019.
\newblock Mathqa: Towards interpretable math word problem solving with
  operation-based formalisms.
\newblock In \emph{Proceedings of the 2019 Conference of the North American
  Chapter of the Association for Computational Linguistics: Human Language
  Technologies, Volume 1 (Long and Short Papers)}, pages 2357--2367.

\bibitem[{Bobrow(1964)}]{Bobrow:1964:NLI:889266}
Daniel~G. Bobrow. 1964.
\newblock Natural language input for a computer problem solving system.
\newblock Technical report, Cambridge, MA, USA.

\bibitem[{Charniak(1969)}]{Charniak:1969:CSC:1624562.1624593}
Eugene Charniak. 1969.
\newblock \href {http://dl.acm.org/citation.cfm?id=1624562.1624593} {Computer
  solution of calculus word problems}.
\newblock In \emph{Proceedings of the 1st International Joint Conference on
  Artificial Intelligence}, IJCAI'69, pages 303--316, San Francisco, CA, USA.
  Morgan Kaufmann Publishers Inc.

\bibitem[{Chen et~al.(2020)Chen, Kornblith, Norouzi, and
  Hinton}]{DBLP:conf/icml/ChenK0H20}
Ting Chen, Simon Kornblith, Mohammad Norouzi, and Geoffrey~E. Hinton. 2020.
\newblock \href {http://proceedings.mlr.press/v119/chen20j.html} {A simple
  framework for contrastive learning of visual representations}.
\newblock In \emph{Proceedings of the 37th International Conference on Machine
  Learning, {ICML} 2020, 13-18 July 2020, Virtual Event}, volume 119 of
  \emph{Proceedings of Machine Learning Research}, pages 1597--1607. {PMLR}.

\bibitem[{Chopra et~al.(2005)Chopra, Hadsell, and LeCun}]{chopra2005learning}
Sumit Chopra, Raia Hadsell, and Yann LeCun. 2005.
\newblock Learning a similarity metric discriminatively, with application to
  face verification.
\newblock In \emph{2005 IEEE Computer Society Conference on Computer Vision and
  Pattern Recognition (CVPR'05)}, volume~1, pages 539--546. IEEE.

\bibitem[{Fang and Xie(2020)}]{DBLP:journals/corr/abs-2005-12766}
Hongchao Fang and Pengtao Xie. 2020.
\newblock \href {http://arxiv.org/abs/2005.12766} {{CERT:} contrastive
  self-supervised learning for language understanding}.
\newblock \emph{CoRR}, abs/2005.12766.

\bibitem[{Gao et~al.(2021)Gao, Yao, and Chen}]{DBLP:conf/emnlp/GaoYC21}
Tianyu Gao, Xingcheng Yao, and Danqi Chen. 2021.
\newblock \href {https://doi.org/10.18653/v1/2021.emnlp-main.552} {Simcse:
  Simple contrastive learning of sentence embeddings}.
\newblock In \emph{Proceedings of the 2021 Conference on Empirical Methods in
  Natural Language Processing, {EMNLP} 2021, Virtual Event / Punta Cana,
  Dominican Republic, 7-11 November, 2021}, pages 6894--6910. Association for
  Computational Linguistics.

\bibitem[{Grill et~al.(2020)Grill, Strub, Altch{\'{e}}, Tallec, Richemond,
  Buchatskaya, Doersch, Pires, Guo, Azar, Piot, Kavukcuoglu, Munos, and
  Valko}]{DBLP:conf/nips/GrillSATRBDPGAP20}
Jean{-}Bastien Grill, Florian Strub, Florent Altch{\'{e}}, Corentin Tallec,
  Pierre~H. Richemond, Elena Buchatskaya, Carl Doersch, Bernardo~{\'{A}}vila
  Pires, Zhaohan Guo, Mohammad~Gheshlaghi Azar, Bilal Piot, Koray Kavukcuoglu,
  R{\'{e}}mi Munos, and Michal Valko. 2020.
\newblock \href
  {https://proceedings.neurips.cc/paper/2020/hash/f3ada80d5c4ee70142b17b8192b2958e-Abstract.html}
  {Bootstrap your own latent - {A} new approach to self-supervised learning}.
\newblock In \emph{Advances in Neural Information Processing Systems 33: Annual
  Conference on Neural Information Processing Systems 2020, NeurIPS 2020,
  December 6-12, 2020, virtual}.

\bibitem[{Guan et~al.(2019)Guan, Liu, Han, Wang, and
  Li}]{guan-etal-2019-improved}
Wenyv Guan, Qianying Liu, Guangzhi Han, Bin Wang, and Sujian Li. 2019.
\newblock \href {https://aclanthology.org/U19-1024} {An improved coarse-to-fine
  method for solving generation tasks}.
\newblock In \emph{Proceedings of the The 17th Annual Workshop of the
  Australasian Language Technology Association}, pages 178--185, Sydney,
  Australia. Australasian Language Technology Association.

\bibitem[{He et~al.(2020)He, Fan, Wu, Xie, and
  Girshick}]{DBLP:conf/cvpr/He0WXG20}
Kaiming He, Haoqi Fan, Yuxin Wu, Saining Xie, and Ross~B. Girshick. 2020.
\newblock \href {https://doi.org/10.1109/CVPR42600.2020.00975} {Momentum
  contrast for unsupervised visual representation learning}.
\newblock In \emph{2020 {IEEE/CVF} Conference on Computer Vision and Pattern
  Recognition, {CVPR} 2020, Seattle, WA, USA, June 13-19, 2020}, pages
  9726--9735. Computer Vision Foundation / {IEEE}.

\bibitem[{Huang et~al.(2016)Huang, Shi, Lin, Yin, and Ma}]{huang2016well}
Danqing Huang, Shuming Shi, Chin-Yew Lin, Jian Yin, and Wei-Ying Ma. 2016.
\newblock How well do computers solve math word problems? large-scale dataset
  construction and evaluation.
\newblock In \emph{Proceedings of the 54th Annual Meeting of the Association
  for Computational Linguistics (Volume 1: Long Papers)}, pages 887--896.

\bibitem[{Huang et~al.(2021)Huang, Wang, Xu, Cao, and Yang}]{huang2021recall}
Shifeng Huang, Jiawei Wang, Jiao Xu, Da~Cao, and Ming Yang. 2021.
\newblock Recall and learn: A memory-augmented solver for math word problems.
\newblock In \emph{Findings of the Association for Computational Linguistics:
  EMNLP 2021}, pages 786--796.

\bibitem[{Johnson et~al.(2019)Johnson, Douze, and
  J{\'e}gou}]{johnson2019billion}
Jeff Johnson, Matthijs Douze, and Herv{\'e} J{\'e}gou. 2019.
\newblock Billion-scale similarity search with {GPUs}.
\newblock \emph{IEEE Transactions on Big Data}, 7(3):535--547.

\bibitem[{Khosla et~al.(2020)Khosla, Teterwak, Wang, Sarna, Tian, Isola,
  Maschinot, Liu, and Krishnan}]{DBLP:conf/nips/KhoslaTWSTIMLK20}
Prannay Khosla, Piotr Teterwak, Chen Wang, Aaron Sarna, Yonglong Tian, Phillip
  Isola, Aaron Maschinot, Ce~Liu, and Dilip Krishnan. 2020.
\newblock \href
  {https://proceedings.neurips.cc/paper/2020/hash/d89a66c7c80a29b1bdbab0f2a1a94af8-Abstract.html}
  {Supervised contrastive learning}.
\newblock In \emph{Advances in Neural Information Processing Systems 33: Annual
  Conference on Neural Information Processing Systems 2020, NeurIPS 2020,
  December 6-12, 2020, virtual}.

\bibitem[{Koncel-Kedziorski et~al.(2015)Koncel-Kedziorski, Hajishirzi,
  Sabharwal, Etzioni, and Ang}]{koncel2015parsing}
Rik Koncel-Kedziorski, Hannaneh Hajishirzi, Ashish Sabharwal, Oren Etzioni, and
  Siena~Dumas Ang. 2015.
\newblock Parsing algebraic word problems into equations.
\newblock \emph{Transactions of the Association for Computational Linguistics},
  3:585--597.

\bibitem[{Koncel-Kedziorski et~al.(2016)Koncel-Kedziorski, Roy, Amini, Kushman,
  and Hajishirzi}]{koncel2016mawps}
Rik Koncel-Kedziorski, Subhro Roy, Aida Amini, Nate Kushman, and Hannaneh
  Hajishirzi. 2016.
\newblock Mawps: A math word problem repository.
\newblock In \emph{Proceedings of the 2016 Conference of the North American
  Chapter of the Association for Computational Linguistics: Human Language
  Technologies}, pages 1152--1157.

\bibitem[{Kumar et~al.(2021)Kumar, Maheshwary, and
  Pudi}]{DBLP:conf/emnlp/KumarMP21}
Vivek Kumar, Rishabh Maheshwary, and Vikram Pudi. 2021.
\newblock \href {https://doi.org/10.18653/v1/2021.findings-emnlp.230}
  {Adversarial examples for evaluating math word problem solvers}.
\newblock In \emph{Findings of the Association for Computational Linguistics:
  {EMNLP} 2021, Virtual Event / Punta Cana, Dominican Republic, 16-20 November,
  2021}, pages 2705--2712. Association for Computational Linguistics.

\bibitem[{Kushman et~al.(2014)Kushman, Artzi, Zettlemoyer, and
  Barzilay}]{kushman2014learning}
Nate Kushman, Yoav Artzi, Luke Zettlemoyer, and Regina Barzilay. 2014.
\newblock Learning to automatically solve algebra word problems.
\newblock In \emph{Proceedings of the 52nd Annual Meeting of the Association
  for Computational Linguistics (Volume 1: Long Papers)}, pages 271--281.

\bibitem[{Li et~al.(2022)Li, Zhang, Yan, Zhou, Li, Liu, and
  Cao}]{li2022seeking}
Zhongli Li, Wenxuan Zhang, Chao Yan, Qingyu Zhou, Chao Li, Hongzhi Liu, and
  Yunbo Cao. 2022.
\newblock Seeking patterns, not just memorizing procedures: Contrastive
  learning for solving math word problems.
\newblock In \emph{Findings of the Association for Computational Linguistics:
  ACL 2022}, pages 2486--2496.

\bibitem[{Liang et~al.(2021)Liang, Zhang, Shao, and Zhang}]{liang2021mwpbert}
Zhenwen Liang, Jipeng Zhang, Jie Shao, and Xiangliang Zhang. 2021.
\newblock \href {http://arxiv.org/abs/2107.13435} {Mwp-bert: A strong baseline
  for math word problems}.

\bibitem[{Liu et~al.(2021)Liu, Guan, Li, Cheng, Kawahara, and
  Kurohashi}]{liu2021roda}
Qianying Liu, Wenyu Guan, Sujian Li, Fei Cheng, Daisuke Kawahara, and Sadao
  Kurohashi. 2021.
\newblock Roda: Reverse operation based data augmentation for solving math word
  problems.
\newblock \emph{IEEE/ACM Transactions on Audio, Speech, and Language
  Processing}, 30:1--11.

\bibitem[{Liu et~al.(2019)Liu, Guan, Li, and Kawahara}]{liu-etal-2019-tree}
Qianying Liu, Wenyv Guan, Sujian Li, and Daisuke Kawahara. 2019.
\newblock \href {https://doi.org/10.18653/v1/D19-1241} {Tree-structured
  decoding for solving math word problems}.
\newblock In \emph{Proceedings of the 2019 Conference on Empirical Methods in
  Natural Language Processing and the 9th International Joint Conference on
  Natural Language Processing (EMNLP-IJCNLP)}, pages 2370--2379, Hong Kong,
  China. Association for Computational Linguistics.

\bibitem[{Miao et~al.(2020)Miao, Liang, and Su}]{DBLP:conf/acl/MiaoLS20}
Shen{-}Yun Miao, Chao{-}Chun Liang, and Keh{-}Yih Su. 2020.
\newblock \href {https://doi.org/10.18653/v1/2020.acl-main.92} {A diverse
  corpus for evaluating and developing english math word problem solvers}.
\newblock In \emph{Proceedings of the 58th Annual Meeting of the Association
  for Computational Linguistics, {ACL} 2020, Online, July 5-10, 2020}, pages
  975--984. Association for Computational Linguistics.

\bibitem[{Patel et~al.(2021)Patel, Bhattamishra, and
  Goyal}]{DBLP:conf/naacl/PatelBG21}
Arkil Patel, Satwik Bhattamishra, and Navin Goyal. 2021.
\newblock \href {https://doi.org/10.18653/v1/2021.naacl-main.168} {Are {NLP}
  models really able to solve simple math word problems?}
\newblock In \emph{Proceedings of the 2021 Conference of the North American
  Chapter of the Association for Computational Linguistics: Human Language
  Technologies, {NAACL-HLT} 2021, Online, June 6-11, 2021}, pages 2080--2094.
  Association for Computational Linguistics.

\bibitem[{Reimers and Gurevych(2019)}]{DBLP:conf/emnlp/ReimersG19}
Nils Reimers and Iryna Gurevych. 2019.
\newblock \href {https://doi.org/10.18653/v1/D19-1410} {Sentence-bert: Sentence
  embeddings using siamese bert-networks}.
\newblock In \emph{Proceedings of the 2019 Conference on Empirical Methods in
  Natural Language Processing and the 9th International Joint Conference on
  Natural Language Processing, {EMNLP-IJCNLP} 2019, Hong Kong, China, November
  3-7, 2019}, pages 3980--3990. Association for Computational Linguistics.

\bibitem[{Roy and Roth(2015)}]{roy-roth-2015-solving}
Subhro Roy and Dan Roth. 2015.
\newblock \href {https://doi.org/10.18653/v1/D15-1202} {Solving general
  arithmetic word problems}.
\newblock In \emph{Proceedings of the 2015 Conference on Empirical Methods in
  Natural Language Processing}, pages 1743--1752, Lisbon, Portugal. Association
  for Computational Linguistics.

\bibitem[{Roy and Roth(2017)}]{roy2017unit}
Subhro Roy and Dan Roth. 2017.
\newblock Unit dependency graph and its application to arithmetic word problem
  solving.
\newblock In \emph{Thirty-First AAAI Conference on Artificial Intelligence}.

\bibitem[{Roy and Roth(2018)}]{roy2018mapping}
Subhro Roy and Dan Roth. 2018.
\newblock Mapping to declarative knowledge for word problem solving.
\newblock \emph{Transactions of the Association of Computational Linguistics},
  6:159--172.

\bibitem[{Shen et~al.(2021)Shen, Yin, Li, Shang, Jiang, Zhang, and
  Liu}]{shen2021generate}
Jianhao Shen, Yichun Yin, Lin Li, Lifeng Shang, Xin Jiang, Ming Zhang, and Qun
  Liu. 2021.
\newblock Generate \& rank: A multi-task framework for math word problems.
\newblock In \emph{Findings of the Association for Computational Linguistics:
  EMNLP 2021}, pages 2269--2279.

\bibitem[{Shen and Jin(2020)}]{shen-jin-2020-solving}
Yibin Shen and Cheqing Jin. 2020.
\newblock \href {https://doi.org/10.18653/v1/2020.coling-main.262} {Solving
  math word problems with multi-encoders and multi-decoders}.
\newblock In \emph{Proceedings of the 28th International Conference on
  Computational Linguistics}, pages 2924--2934, Barcelona, Spain (Online).
  International Committee on Computational Linguistics.

\bibitem[{Shen et~al.(2022)Shen, Liu, Mao, Wan, Cheng, and
  Kurohashi}]{shen2022seeking}
Yibin Shen, Qianying Liu, Zhuoyuan Mao, Zhen Wan, Fei Cheng, and Sadao
  Kurohashi. 2022.
\newblock Seeking diverse reasoning logic: Controlled equation expression
  generation for solving math word problems.
\newblock \emph{arXiv preprint arXiv:2209.10310}.

\bibitem[{Shi et~al.(2015)Shi, Wang, Lin, Liu, and Rui}]{shi2015automatically}
Shuming Shi, Yuehui Wang, Chin-Yew Lin, Xiaojiang Liu, and Yong Rui. 2015.
\newblock Automatically solving number word problems by semantic parsing and
  reasoning.
\newblock In \emph{Proceedings of the 2015 Conference on Empirical Methods in
  Natural Language Processing}, pages 1132--1142.

\bibitem[{Tan et~al.(2021)Tan, Wang, Jiang, and Jiang}]{tan2021investigating}
Minghuan Tan, Lei Wang, Lingxiao Jiang, and Jing Jiang. 2021.
\newblock \href {http://arxiv.org/abs/2105.08928} {Investigating math word
  problems using pretrained multilingual language models}.

\bibitem[{Upadhyay and Chang(2017)}]{upadhyay2017annotating}
Shyam Upadhyay and Ming-Wei Chang. 2017.
\newblock Annotating derivations: A new evaluation strategy and dataset for
  algebra word problems.
\newblock In \emph{Proceedings of the 15th Conference of the European Chapter
  of the Association for Computational Linguistics: Volume 1, Long Papers},
  pages 494--504.

\bibitem[{Wang et~al.(2018)Wang, Zhang, Gao, Song, Guo, and
  Shen}]{wang2018mathdqn}
Lei Wang, Dongxiang Zhang, Lianli Gao, Jingkuan Song, Long Guo, and Heng~Tao
  Shen. 2018.
\newblock Mathdqn: Solving arithmetic word problems via deep reinforcement
  learning.
\newblock In \emph{Proceedings of the AAAI Conference on Artificial
  Intelligence}, volume~32.

\bibitem[{Wang et~al.(2019)Wang, Zhang, Zhang, Xu, Gao, Dai, and
  Shen}]{wang2019template}
Lei Wang, Dongxiang Zhang, Jipeng Zhang, Xing Xu, Lianli Gao, Bing~Tian Dai,
  and Heng~Tao Shen. 2019.
\newblock Template-based math word problem solvers with recursive neural
  networks.
\newblock In \emph{Proceedings of the AAAI Conference on Artificial
  Intelligence}, volume~33, pages 7144--7151.

\bibitem[{Wang et~al.(2017)Wang, Liu, and Shi}]{DBLP:conf/emnlp/WangLS17}
Yan Wang, Xiaojiang Liu, and Shuming Shi. 2017.
\newblock \href {https://doi.org/10.18653/v1/d17-1088} {Deep neural solver for
  math word problems}.
\newblock In \emph{Proceedings of the 2017 Conference on Empirical Methods in
  Natural Language Processing, {EMNLP} 2017, Copenhagen, Denmark, September
  9-11, 2017}, pages 845--854. Association for Computational Linguistics.

\bibitem[{Xie and Sun(2019)}]{xie2019goal}
Zhipeng Xie and Shichao Sun. 2019.
\newblock A goal-driven tree-structured neural model for math word problems.
\newblock In \emph{IJCAI}, pages 5299--5305.

\bibitem[{Zhang et~al.(2020{\natexlab{a}})Zhang, Lee, Lim, Qin, Wang, Shao, and
  Sun}]{ijcai2020-555}
Jipeng Zhang, Roy Ka-Wei Lee, Ee-Peng Lim, Wei Qin, Lei Wang, Jie Shao, and
  Qianru Sun. 2020{\natexlab{a}}.
\newblock \href {https://doi.org/10.24963/ijcai.2020/555} {Teacher-student
  networks with multiple decoders for solving math word problem}.
\newblock In \emph{Proceedings of the Twenty-Ninth International Joint
  Conference on Artificial Intelligence, {IJCAI-20}}, pages 4011--4017.
  International Joint Conferences on Artificial Intelligence Organization.
\newblock Main track.

\bibitem[{Zhang et~al.(2020{\natexlab{b}})Zhang, Wang, Lee, Bin, Wang, Shao,
  and Lim}]{zhang2020graph}
Jipeng Zhang, Lei Wang, Roy Ka-Wei Lee, Yi~Bin, Yan Wang, Jie Shao, and Ee-Peng
  Lim. 2020{\natexlab{b}}.
\newblock Graph-to-tree learning for solving math word problems.
\newblock In \emph{Proceedings of the 58th Annual Meeting of the Association
  for Computational Linguistics}, pages 3928--3937.

\bibitem[{Zou and Lu(2019)}]{zou2019text2math}
Yanyan Zou and Wei Lu. 2019.
\newblock Text2math: End-to-end parsing text into math expressions.
\newblock In \emph{Proceedings of the 2019 Conference on Empirical Methods in
  Natural Language Processing and the 9th International Joint Conference on
  Natural Language Processing (EMNLP-IJCNLP)}, pages 5330--5340.

\end{thebibliography}
\bibliographystyle{acl_natbib}

\end{document}